\documentclass{article}

\PassOptionsToPackage{numbers, compress}{natbib}

\usepackage[final]{neurips_data_2021}

\usepackage[utf8]{inputenc} 
\usepackage[T1]{fontenc}    
\usepackage{hyperref}       
\usepackage{url}            
\usepackage{booktabs}       
\usepackage{amsfonts}       
\usepackage{nicefrac}       
\usepackage{microtype}      
\usepackage{xcolor}         
\usepackage{comment}        
\usepackage{array}          
\newcolumntype{M}[1]{>{\centering\arraybackslash}m{#1}}
\usepackage{caption}        
\usepackage{tabularx}       
\usepackage{makecell}       
\usepackage{longtable}      
\usepackage{listings}       
\usepackage{amsmath}
\usepackage{graphicx}
\usepackage{longtable}
\newboolean{include-notes}
\setboolean{include-notes}{true}

\newcommand{\authorsep}{\vspace{-.3em}}

\newcommand{\TODO}[1]{\ifthenelse{\boolean{include-notes}}
 {{\color{red} TODO: #1}}{}}
\newcommand{\ELTODO}[1]{\ifthenelse{\boolean{include-notes}}
 {{\color{orange} ELTODO: #1}}{}}
 \newcommand{\NATODO}[1]{\ifthenelse{\boolean{include-notes}}
 {{\color{blue} NATODO: #1}}{}}
\newcommand{\name}{\textsc{RAFT}}

\newcommand{\benchmarkURL}{\href{https://raft.elicit.org}{https://raft.elicit.org}}
\newcommand{\submissionInstructionsURL}{\href{https://raft.elicit.org/submit}{https://raft.elicit.org/submit}}

\newcommand{\baselinesURL}{\href{https://raft.elicit.org/baselines}{https://raft.elicit.org/baselines}}
\newcommand{\datasetsURL}{\href{https://raft.elicit.org/datasets}{https://raft.elicit.org/datasets}}

\title{RAFT: A Real-World Few-Shot \\ Text Classification Benchmark}

\author{%
  Neel Alex\thanks{Equal contribution. Correspondence to eli.d.lifland@gmail.com and salexucb@berkeley.edu. NA contributed during an internship at Ought.} \authorsep \\
  Ought \authorsep \\
  \And
  Eli Lifland\footnotemark[1] \authorsep \\
  Ought \authorsep \\
  \And
  Lewis Tunstall \authorsep \\
  Hugging Face \authorsep \\
  \AND
  Abhishek Thakur \authorsep \\
  Hugging Face \authorsep \\
  \And
  Pegah Maham \authorsep \\
  Stiftung Neue Verantwortung \authorsep \\  
  \And
  C. Jess Riedel \authorsep \\
  NTT Research \authorsep \\
  \AND
  Emmie Hine \authorsep \\
  Oxford Internet Institute \authorsep \\
  \And
  Carolyn Ashurst \authorsep \\
  Alan Turing Institute \authorsep \\
  \And
  Paul Sedille \authorsep \\
  Harvard Kennedy School \authorsep \\      
  \AND
  Alexis Carlier \authorsep \\
  University of Oxford \authorsep \\    
  \And
  Michael Noetel \authorsep \\
  Australian Catholic University \authorsep \\  
  \And
  Andreas Stuhlmüller \authorsep \\
  Ought \authorsep \\
  }

\begin{document}

\maketitle

\begin{abstract}
Large pre-trained language models have shown promise for few-shot learning, completing text-based tasks given only a few task-specific examples. Will models soon solve classification tasks that have so far been reserved for human research assistants? Existing benchmarks are not designed to measure progress in applied settings, and so don't directly answer this question. The \name{} benchmark (\textbf{R}eal-world \textbf{A}nnotated \textbf{F}ew-shot \textbf{T}asks) focuses on naturally occurring tasks and uses an evaluation setup that mirrors deployment. Baseline evaluations on \name{} reveal areas current techniques struggle with: reasoning over long texts and tasks with many classes. Human baselines show that some classification tasks are difficult for non-expert humans, reflecting that real-world value sometimes depends on domain expertise. Yet even non-expert human baseline F1 scores exceed GPT-3 by an average of $0.11$. The \name{} datasets and leaderboard will track which model improvements translate into real-world benefits at \benchmarkURL{}.
\end{abstract}

\section{Introduction}

\begin{figure}
\includegraphics[width=\textwidth]{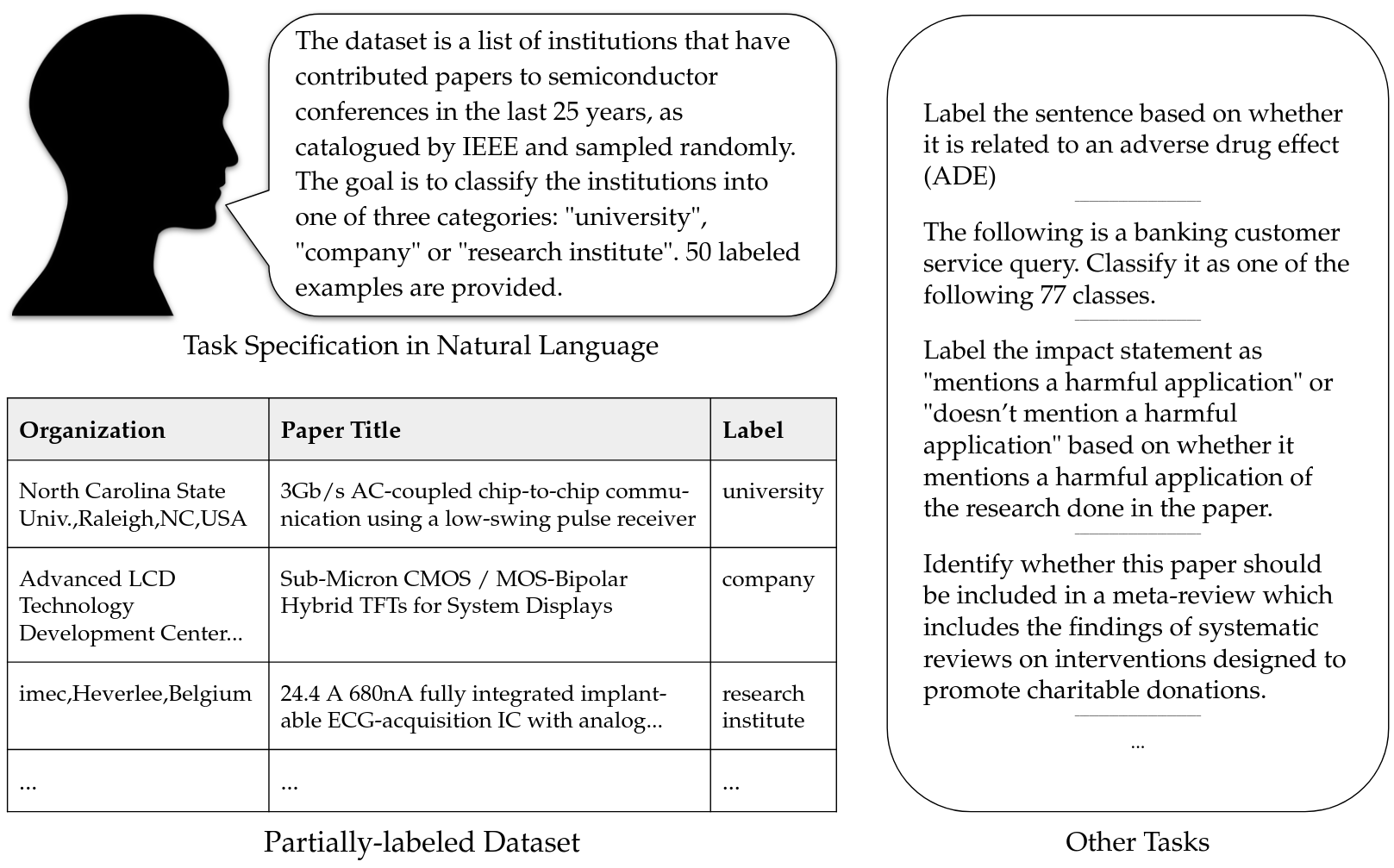}
\centering
\caption{\name{} includes naturally occurring classification datasets, mimicking work that is usually given to human research assistants. Each task comes with natural language instructions and labels in addition to 50 training examples.}
\end{figure}

Few-shot learning, the capacity to complete a task given a small number of demonstrations \citep{FeiFei2006OneshotLO}, is one of the hallmarks of human intelligence \citep{Tenenbaum2011HowTG, Lake2011OneSL}. As researchers, we leverage this capacity when we delegate work on crowdsourcing platforms or give a task with examples to a human research assistant.

\citet{brown2020gpt3} show that large pre-trained language models exhibit few-shot learning capabilities for a wide range of natural language tasks.
If those capabilities were comparable to people on economically relevant tasks, this would be important to know: a single model could be used across multiple real-world tasks, with low per-task data labeling cost.
However, these models have also been shown to have inconsistent few-shot performance depending on the exact setup and task being solved \citep[e.g.][]{mishra2021natural, perez2021true}.
The mixed evidence suggests that it would be valuable to measure and track few-shot performance on a set of tasks that is representative of what appears in practice.

Natural language tasks coarsely split into generation, classification, and retrieval. We focus on classification tasks because they support high-quality automated evaluation, cover a wide range of economically valuable tasks, and yet don't have existing real-world benchmarks.

Existing few-shot classification benchmarks are typically designed to highlight areas where models fall short \citep{schick2021its} or to study particular model abilities \citep{bragg2021flex, ye2021crossfit, mishra2021natural}. The tasks and evaluation setup aren't optimized to measure progress in applied settings:

\begin{itemize}

\item \textit{Tasks} that are generated or chosen specifically to test language models may not represent some of the challenges found when applying these models in real-world settings. For example, SuperGLUE \citep{wang2019superglue} and the few-shot equivalent FewGLUE \citep{schick2021its} mainly include short texts. Doing well on applied tasks sometimes requires reasoning over long texts. Existing systems struggle with long texts due to a limited context window, especially in the few-shot setting where some systems learn from examples presented in context.

\item The \textit{evaluation} does not closely mirror deployment, and may both under- and overestimate models' capabilities.
It may underestimate model capability by restricting models to the closed-book setting (e.g., no retrieval from online sources) and using uninformative labels (e.g., 0/1 instead of ``about literature'' vs.\ ``about movies'').
It may overestimate model capability by using many more than a few examples for setting hyperparameters during validation \citep{perez2021true}.

\end{itemize}

\name{} is a real-world few-shot text classification benchmark designed to measure how much recent and upcoming NLP advances benefit applications:

\begin{itemize}

\item The \textit{tasks} are naturally occurring tasks. Their labeling is inherently valuable to someone, and they may have challenges that are not reflected in synthetic tasks. Inherent value means that, if it were sufficiently fast and cheap, it would be desirable to outsource the task to human research assistants or crowd workers. Challenges refers to the need for information retrieval, domain expertise, parsing long documents, and making use of instructions. Table \ref{tab:basic-dataset} shows the real-world challenges presented by \name{}, including 4 datasets with long input texts.

\item The \textit{evaluation} closely mirrors deployment. For each task, we release a public training set with 50 examples and a larger unlabeled test set\footnote{Datasets are at \datasetsURL{}}. We encourage unsupervised pre-training on the unlabelled examples and open-domain information retrieval. We keep the test-set labels private and provide automated evaluation through a Hugging Face leaderboard\footnote{Instructions for submission are at \submissionInstructionsURL}.

\end{itemize}

In addition to the gold-standard human labels, we collect automatic and crowdsourced baselines. The automatic baselines reveal areas where current techniques struggle, such as reasoning over long texts and tasks with many classes. The crowdsourced baseline reveals that \name{} includes a mix of moderate to difficult tasks. We also observe difficulties in collecting human crowdsourced baselines on some datasets, particularly when domain expertise is important, which suggests that real-world value often depends on domain knowledge.

The \name{} datasets and leaderboard can be viewed and submitted to at \benchmarkURL{}.

\section{Related Work}

We briefly review few-shot learning in NLP, then the benchmarks that are most similar to \name{}.

\subsection{Few-shot learning in NLP}

Pre-trained language models (PLMs) such as BERT \citep{devlin2019bert} and GPT-3 \citep{brown2020gpt3} can learn to do some NLP tasks when prompted with a few demonstrations, including some classification tasks. The two primary approaches to few-shot classification using PLMs are in-context learning and prompt-based fine-tuning.

\textbf{In-context learning.} A PLM is primed with labeled examples in its prompt. It classifies the example included at the end of the prompt by predicting the classification conditioned on the priming. GPT-3 \citep{brown2020gpt3} used in-context learning to achieve promising results on a variety of classification tasks. UniFew \citep{bragg2021flex} similarly achieved strong results on classification tasks via in-context learning, converting classification tasks into a multiple-choice question answer format for prompting.

\textbf{Prompt-based fine-tuning}. A PLM is fine-tuned with masked-language modeling objectives to learn from few examples. This is also known as Pattern-exploiting training (PET) \citep{schick2021exploiting}. While PET requires task-specific prompts, it achieves better performance than GPT-3 in-context with smaller models \citep{schick2021its}. LM-BFF \citep{gao2021making} improves prompt-based fine-tuning by dynamically constructing prompts.

\subsection{Few-shot NLP benchmarks}

The most closely related few-shot NLP benchmarks are FLEX \cite{bragg2021flex}, FewGLUE \cite{schick2021its}, CrossFit \cite{ye2021crossfit}, and NaturalInstructions \cite{mishra2021natural}. Each of these benchmarks includes at least some classification tasks with meaningful textual labels.

These benchmarks are designed to study transfer between tasks \citep{bragg2021flex, ye2021crossfit}, pinpoint where NLP models fall short \citep{schick2021its}, and evaluate ability of models to follow instructions \citep{mishra2021natural}, whereas \name{} is designed to be representative of real-world classification tasks. This difference in goals is reflected in selection of tasks and evaluation:


\textbf{Tasks.} FLEX, FewGLUE and NaturalInstructions test on traditional NLP tasks. CrossFit tests on 160 tasks from the Hugging Face Datasets\footnote{\href{https://huggingface.co/datasets}{https://huggingface.co/datasets}} and includes some naturally occurring datasets, including the TweetEval dataset \citep{barbieri-etal-2020-tweeteval} that \name{} uses as well. CrossFit excludes tasks that leverage external sources or information retrieval techniques, need domain knowledge (e.g. COVID-19 datasets), and long documents (e.g. scientific papers). Like \name{}, FLEX includes tasks with strong class imbalance. 

\textbf{Evaluation.} None of the existing benchmarks allow open-domain information retrieval. Like FLEX, \name{} provides no extra validation data beyond the training examples. \citet{perez2021true} argue that the performance of state-of-the-art few-shot methods has been overestimated by most other existing benchmarks because they use labeled examples beyond the few training instances provided for model and parameter selection.

\section{Benchmark Description}

\name{} is a few-shot {\em classification} benchmark. We focus on classification primarily because automatic evaluation is more reliable than for generation tasks. We believe (as our results will later confirm) that there still is a substantial gap between even non-expert humans and automated systems in the few-shot classification setting.

Both tasks (datasets and metadata) and evaluation (rules for submission, metrics) are chosen to mirror real-world classification problems.

\subsection{Tasks}

A classification task is a dataset with labeled natural language entries. Each label corresponds one-to-one with a natural language class name. Each task has instructions for labeling.

\subsubsection{Dataset selection criteria}
\label{sec:dataset-selection}

We selected datasets based on the following criteria (``non-trivial real world tasks''):

\textbf{Naturally occurring.} We focus on data that are naturally occurring, rather than being synthetically generated to test and improve language models.

\textbf{Intrinsic value.} We select datasets for which the correct labeling inherently provides real-world value.  \name{} includes tasks like hate-speech detection, medical case report parsing, and literature review automation, where better performance translates into practical benefits. This criterion involves subjectivity, but we aimed to select tasks that approximate the distribution of valuable classification tasks well.

\textbf{Realistic class distribution.} We did not exclude datasets with heavily imbalanced classes.

\textbf{Open-domain feasibility.} As we provide an open-domain setting where information retrieved from the web may be used to augment predictions, we excluded tasks for which the correct label is extremely easily discoverable through a Google search. For example, we considered including the LIAR \citep{wang-2017-liar} dataset which includes Politifact statements and their veracity. We decided against including it since it would be trivial to get 100\% accuracy by running a site search on \href{https://www.politifact.com/}{https://www.politifact.com/}.

In order to gather datasets meeting the above requirements, we put out a collaboration request. We also reached out to users of classification on Elicit \citep{elicit}. Lastly, we conducted a search of existing datasets on the Hugging Face Hub\footnote{\href{https://huggingface.co/datasets}{https://huggingface.co/datasets}} and PapersWithCode \footnote{\href{https://paperswithcode.com}{https://paperswithcode.com}}.

\subsubsection{Dataset preparation}

In cases where the test set was over 5,000 data points, we randomly selected 5,000 to serve as a test set in order to keep the test set sizes manageable. When the dataset didn't already have textual labels, we added textual labels according to our best understanding of the task.

\subsubsection{Selected \name{} datasets}
\label{sec:dataset-description}

\begin{table}[ht]
\small
\centering
\begin{tabular}{m{15em}M{4.5em}M{4.5em}M{5em}M{4.5em}M{4.5em}}
\toprule
\textbf{Dataset Name} & \textit{Long inputs} & \textit{Domain expertise} & \textit{Detailed instructions} & \textit{Number of classes} & \textit{Test set size} \\
\midrule
ADE Corpus V2 (\textit{ADE}) & -- & \checkmark & -- & 2 & 5,000 \\ 
Banking77 (\textit{B77}) & -- & -- & --  & 77 & 5,000 \\    
NeurIPS impact statement risks (\textit{NIS}) & \checkmark & -- & -- & 2 & 150 \\ 
OneStopEnglish (\textit{OSE}) & \checkmark & -- & -- & 3 & 516  \\ 
Overruling (\textit{Over}) & -- & \checkmark & -- & 2 & 2,350 \\ 
Semiconductor org types (\textit{SOT}) & -- & -- & -- & 3 & 449 \\ 
Systematic review inclusion (\textit{SRI}) & \checkmark & -- & \checkmark & 2 & 2,243 \\ 
TAI safety research (\textit{TAI}) & \checkmark & \checkmark & \checkmark & 2 & 1,639 \\ 
Terms of Service (\textit{ToS}) & -- & \checkmark & \checkmark & 2 & 5,000 \\ 
TweetEval Hate (\textit{TEH}) & -- & -- & \checkmark & 2 & 2,966 \\ 
Twitter complaints (\textit{TC}) & -- & -- & -- & 2 & 3,399 \\ 
\bottomrule
\end{tabular}
\caption{Overview of the tasks in \name{}. \textit{Long inputs}, \textit{Domain expertise}, and \textit{Detailed instructions} are some of the real-world challenges posed by \name{}.}
\label{tab:basic-dataset}
\end{table}

We selected 11 datasets, accessible at \datasetsURL{}. Table \ref{tab:basic-dataset} presents an overview of the datasets. More details are available in the Appendix.

\textbf{ADE Corpus V2 (\textit{ADE}).} The ADE corpus V2 \citep{GURULINGAPPA2012885} contains sentences from medical case reports annotated for relation to adverse drug effects. We focus on the binary classification task of whether a sentence is related to an adverse drug effect (ADE).


\textbf{Banking77 (\textit{B77}).} Banking77  \citep{casanueva2020efficient} contains online banking customer service queries annotated with their intents.


\textbf{NeurIPS impact statement risks (\textit{NIS}).} We include the broader impact statements from NeurIPS 2020 papers collected in the dataset from \citet{ashurst2021aiethics}. We annotate these based on whether they mention possibly harmful applications of the research done in the paper .\footnote{The raw scraped NeurIPS impact statements can be found at \href{https://raft.elicit.org/neurips-impact}{https://raft.elicit.org/neurips-impact}.}

\textbf{OneStopEnglish (\textit{OSE}).} OneStopEnglish \citep{vajjala-lucic-2018-onestopenglish} contains articles sourced from The Guardian newspaper and rewritten by teachers to suit three levels of adult English as a Second Language (ESL) learners.

\textbf{Overruling (\textit{Over}).} Overruling \citep{zheng2021does} contains statements from a law corpus annotated based on whether they are overruling, defined as nullifying a previous case decision as a precedent.

\textbf{Semiconductor org types (\textit{SOT}).} We collect a dataset of institutions that have contributed to semiconductor conferences in the last 25 years, then classify these institutions into organization types: ``university", ``company", and ``research institute".

\textbf{Systematic review inclusion (\textit{SRI}).} We use data from a systematic meta-review studying interventions to increase charitable donations \citep{noetel_slattery_saeri_lee_houlden_farr_gelber_stone_huuskes_timmons_2020}. The task is to predict whether a paper advances past the screening stage.

\textbf{TAI safety research (\textit{TAI}).} We include data from the formation of a bibliographic database for research on the safety of transformative artificial intelligence (TAI) \citep{riedel_deibel_2020}. We choose the binary task of predicting whether a work is classified as TAI safety research.

\textbf{Terms of Service (\textit{ToS}).} The Terms of Service dataset \citep{Lippi_2019} contains clauses from Terms of Services, annotated by whether they are potentially unfair to consumers.

\textbf{TweetEval Hate (\textit{TEH}).} We include the hate-speech detection task from the TweetEval dataset \citep{barbieri-etal-2020-tweeteval}, which was curated from \citet{basile-etal-2019-semeval}.

\textbf{Twitter complaints (\textit{TC}).} We include a dataset of tweets annotated by whether they contain a complaint \citep{preotiuc-pietro-etal-2019-automatically}.

\subsection{Evaluation}

\subsubsection{Setting and rules}

The \name{} evaluation replicates real-world few-shot classification problems by restricting to 50 labeled examples without validation set, providing meaningful instructions and labels, and using a no-holds-barred setting:

\textbf{50 labeled examples.} We provide 50 labeled examples per task (not per class). In the authors' experience with users of the classification tool Elicit \citep{elicit}, this is approximately the number of examples people are willing to label for a task with a few thousand unlabeled examples. The 50 examples are chosen randomly, mirroring the applied setting in which one can't easily choose a balanced set. No examples beyond the chosen 50 are available for validation.\footnote{By only releasing 50 labelled examples, we make it difficult to cheat by using more than 50 examples for validation. For the NeurIPS impact statement risks and Semiconductor org type datasets, the test set labels aren't available publicly. For other datasets, the test set labels are available publicly but it is non-trivial and discouraged to seek them out.}

\textbf{Task-specific instructions.} As an important replacement for large amounts of labeled data, instructions can specify how a task should be done. Therefore, we provide the instructions we give to human labelers so that they can be used in instructing automatic systems. The level of detail of the instructions varies. We write the instructions based on information from publications (for datasets published elsewhere) or in consultation with the dataset creator (for new datasets).

\textbf{Meaningful label names.} Similar to instructions, textual labels are an important aspect of few-shot and especially zero-shot learning. We create default textual labels for each dataset as recommended by FLEX \citep{bragg2021flex}.

\textbf{Transfer learning permitted.} Transfer and meta-learning using other datasets is permitted, including further pre-training on other corpora.

\textbf{Unlabeled data permitted.} Use of the unlabeled \name{} test sets is permitted, as unlabeled data are usually available in the applied setting.

\textbf{Open-domain retrieval permitted.} Models may be augmented with information retrieved from the internet, e.g. via automated web searches.\footnote{Ideally, we'd only allow information from before the time each dataset needed to be labeled. This isn't currently feasible, so we settle for a fully open-domain setting.}

\textbf{Submission requires only labels.} Submission on the test set is open to the public and only requires upload of test set labels. This is in line with benchmarks like GLUE \citep{wang2019glue} and SuperGLUE \citep{wang2019superglue}, but is in contrast to the few-shot benchmark FLEX \citep{bragg2021flex}. By only requiring labels, we give submission creators maximal flexibility in what models to set up.

\textbf{Weekly evaluation.} Evaluation is run on a weekly basis to minimize information gained from frequent repeated submissions.

\begin{table}[ht]
\small
\centering
\begin{tabular}{m{9.5em} M{1.5em} M{1.5em} M{1.5em} M{1.5em} M{1.5em} M{1.5em} M{1.5em} M{1.5em} M{1.5em} M{1.5em} M{1.5em} M{1.5em} M{1.5em}}
\toprule
\textbf{Baseline} & \textbf{Avg} & \textit{ADE} & \textit{B77} & \textit{NIS} & \textit{OSE} & \textit{Over} & \textit{SOT} & \textit{SRI} & \textit{TAI} & \textit{ToS} & \textit{TEH} & \textit{TC} \\  
\midrule
Human (crowdsourced) & \textbf{.735} & \textbf{.830} & \textbf{.607} & \textbf{.857} & \textbf{.646} & .917 & \textbf{.908} & .468 & .609 & \textbf{.627} & \textbf{.722} & \textbf{.897} \\
GPT-3 (175B) & .627 & .686 & .299 & .679 & .431 & \textbf{.937} & .769 & \textbf{.516} & \textbf{.656} & .574 & .526 & .821 \\
AdaBoost & .514 & .543 & .023 & .626 & .475 & .838 & .455 & .506 & .556 & .560 & .443 & .625 \\
GPT-Neo (2.7B) & .481 & .452 & .149 & .408 & .343 & .681 & .406 & .493 & .605 & .565 & .554 & .636 \\
GPT-2 (1.6B) & .458 & .600 & .121 & .561 & .245 & .498 & .380 & .492 & .612 & .498 & .311 & .723 \\
BART MNLI Zero-shot & .382 & .234 & .332 & .615 & .360 & .462 & .644 & .026 & .469 & .122 & .543 & .400 \\
Plurality class & .331 & .446 & .000 & .353 & .164 & .337 & .271 & .493 & .344 & .471 & .366 & .391 \\
GPT-3 Zero-shot & .292 & .163 & .000 & .572 & .323 & .378 & .628 & .027 & .362 & .164 & .303 & .290 \\
\bottomrule
\end{tabular}
\caption{Performance of \name{} baselines (F1)}
\label{tab:baseline-results}
\end{table}

\subsection{Metrics}

Since some \name{} datasets have substantial class imbalances, we use F1 as our evaluation metric. We compute macro-averaged F1 scores, even for binary datasets. To get an overall score, we average across all datasets.

\section{Baselines}

The code for all automatic baselines is open-sourced at \baselinesURL{}.

\subsection{GPT-3 baseline}

We provide a simple automatic baseline using GPT-3 \citep{brown2020gpt3}, accessed through the OpenAI API\footnote{\href{https://beta.openai.com}{https://beta.openai.com}}. As in \citet{brown2020gpt3}, we use in-context learning, adding labeled examples to the prompt to prime GPT-3. We also run a zero-shot version with no training examples included in the prompt.

\subsubsection{Prompt construction}

We build a prompt consisting of:
\begin{enumerate}
    \item Task-specific instructions
    \item $N$ labeled examples, with $N$ selected on a per-task basis
    \item The target example to classify
\end{enumerate}

Example prompts for all datasets are available at \href{https://raft.elicit.org/baseline-prompts}{https://raft.elicit.org/baseline-prompts}.

\textbf{Truncation.} GPT-3's context window support up to $2,048$ tokens. The experiments in \citet{brown2020gpt3} include as many complete labeled examples as would fit in the context, reporting that typically 10 to 100 examples fit. However, datasets such as \textit{OSE} have very long inputs so that only 1-2 complete labeled examples would fit, suggesting that another approach may be better. 

We select $N$ training examples to include in a given prompt and then truncate the examples. In a given task, the instructions take up $I$ tokens. Separators between the instructions and each example take up $S$ tokens. No more than $T = 2,048$ tokens can be used. 
\begin{enumerate}
    \item $E = T - I - S$ tokens are allotted for the training examples and classification target.
    \item The classification target is truncated to $\frac{1}{4} E$ tokens. 
    \item Each of the $N$ remaining training examples is truncated to $\frac{3}{4}\frac{E}{N}$ tokens. 
\end{enumerate}

We truncate from a training example's data fields first, leaving the label intact.

\textbf{Field selection and sorting.} We exclude data fields that are unlikely to contribute substantially to GPT-3's performance. These fields either deal with the authors of the textual example or are URLs. Additionally, we sort the order in which the text fields occur to put the most important fields first. When examples are truncated, the most important information is preserved. 

\textbf{Semantic selection.} To select training examples to include in the prompt for a given test example, we selected the most similar training examples as in \citet{liu2021makes}. To perform semantic search, we use the OpenAI API search endpoint with the \texttt{ada} engine.

\subsubsection{Classification}

With the prompt formed, we retrieve GPT-3's 100 most likely next tokens using the \texttt{davinci} engine. For each class, we assign the probability that its first token is generated. We then normalize the probabilities to sum to 1. For the \textit{B77} dataset, multiple labels share the same first token so we prepend a numerical prefix such as ``1. '' to each class.

\subsubsection{Parameter selection}
\label{sec:param-tuning}

We tune the GPT-3 baseline on the training set using \textit{leave-one-out cross validation} (LOOCV): $k$-fold cross validation with $k = n$ so that only one test example is used at a time for validation. While LOOCV isn't robust with as few as 50 examples as discussed in \citet{perez2021true}, it is one of the best options for parameter selection in the few-shot setting. Detailed LOOCV results are in Section \ref{sec:auto-baseline-details}.

\textbf{Instructions.} We test two modes of instruction: (a) a generic classification prompt: "Possible labels:" followed by a list of textual labels. (b) instructions similar to the ones given to human labelers, plus the list of textual labels. The instructions are taken whole when possible, and otherwise shortened and summarized manually to limit usage of the GPT-3 context window. Task-specific instructions outperform generic instructions by an $.04$ on averaged F1 score, thus we include task-specific instructions in the baseline.

\textbf{Semantic training example selection.} To select training examples for inclusion in the prompt from a larger set, we consider (a) selecting examples randomly and (b) using semantic search to identify the training examples most similar to the test example. Semantic selection outperforms random selection by $0.03$ on averaged F1, thus we include semantic selection in the baseline.

\textbf{Number of examples in the prompt.} We select the number of examples to include in the prompt on a per-dataset basis, as our truncation strategy induces a quality-quantity trade-off. For each dataset, we test performance with 5, 10, 25, and 50\footnote{49 rather than 50 training examples for LOO experiments} training examples and choose the number that performs best by F1. For datasets with long inputs, smaller numbers of more detailed samples often produce better performance, while datasets with smaller inputs can fit more complete labeled examples in the prompt.

\subsection{Other automatic baselines}

\textbf{In-context baselines.} We run further in-context baselines GPT-Neo \citep{gpt-neo} and GPT-2 \citep{radford2019language}. We provide code\footnote{\baselinesURL{}} for generating predictions on \name{} using these models and any other causal language model available on the HuggingFace Hub. For semantic search, we use a MiniLM \citep{wang2020minilm} fine-tuned on sentence pairs via the sentence-transformers package\footnote{https://huggingface.co/sentence-transformers/all-MiniLM-L6-v2}.

\textbf{Zero-shot baselines.} We run two transformers in the zero-shot setting: 
\begin{itemize}
    \item GPT-3, to judge to what extent training examples in the prompt aid performance
    \item BART \citep{bart2019lewis} trained on MNLI \citep{mnli2017williams}, as suggested by \citet{zeroshot2019yin} and \citet{davison2020zeroshotlearning} as an effective zero-shot classification approach
\end{itemize}

\textbf{Non-neural baselines.} We run AdaBoost \citep{FREUND1997119} to establish a strong non-neural baseline. We construct feature vectors for each example based on the counts of $n$-grams of $1$-$5$ words as the input to a weighted ensemble of 100 depth-3 decision trees. These decision trees and weights are trained with AdaBoost with learning rate 1, and evaluated through weighted voting. We also include a plurality (most frequent) class baseline.

\subsection{Human baseline}

To collect human baselines, we use the Surge\footnote{https://www.surgehq.ai/} crowdsourcing platform. Following \citet{wang2019superglue}, we randomly select 100 data points from each test set and use a 2-step labeling process: qualification then annotation. The crowdsourced label is the plurality vote of 5 labelers.

We put crowd workers in a similar situation to automated systems. We link to a sheet with the same 50 labeled examples, use the same textual labels, and give the same task-specific instructions that we are providing to practitioners to adapt for instructing language models.\footnote{For details on the human baseline gathering process, see Section \ref{sec:human-baseline-details}.}

\subsection{Analysis}

\textbf{Humans generally outperform GPT-3.} Humans outperform GPT-3 on 8 out of 11 tasks, demonstrating room for improvement for models on real-world few-shot tasks. We expect that exceeding the crowdsourced baseline will require substantial advances in model performance, and even more so for a future expert human baseline.

Weaknesses of GPT-3 include:

\begin{itemize}
    \item \textbf{Many classes}: Humans most outperform GPT-3 on \textit{B77}, which has by far the most classes in \name{}. With 77 classes and 50 labeled examples, many classes have no corresponding labeled examples. Additionally, just listing out the possible classes takes up a large portion of GPT-3's context window.
    \item \textbf{Long inputs}: GPT-3 performs poorly on some tasks requiring reasoning over long inputs, such as \textit{NIS} and \textit{OSE}. GPT-3's context window may be a contributing factor.
\end{itemize}

\textbf{Crowd-sourced baselines struggle on domain-specific tasks.} 
Crowd-sourced humans substantially outperform GPT-3 on only 1 of 4 tasks we identified as requiring domain expertise:

\begin{itemize}
    \item Humans substantially outperform GPT-3 on \textit{ADE}, which requires medical expertise.
    \item Humans outperformed GPT-3 by just .053 on \textit{ToS}, which requires parsing legal language.
    \item GPT-3 outperforms humans on \textit{Over}, which requires greater legal expertise than \textit{ToS} \citep{zheng2021does}, and \textit{TAI}, which requires expertise in AI safety research.
\end{itemize}

\textbf{Zero-shot performance is weak.} GPT-3 zero-shot does poorly on \name{}, performing worse than the plurality class baseline. BART zero-shot exceeds the plurality class baseline but does not do so in every dataset, and it is not competitive with few-shot language models. We encourage future research on improving performance in the zero-shot setting, perhaps through improved prompt construction and transfer learning.

\textbf{Neural baselines besides few-shot GPT-3 perform worse than AdaBoost.} Generative language models smaller than GPT-3 comfortably outperform the plurality class baseline but remain below AdaBoost. We use the same amount of labelled examples in the prompt as with GPT-3 despite the context window being smaller; performance may improve with fewer (but longer) examples.

\section{Discussion}

\subsection{Limitations}

\textbf{Linguistic diversity.} The benchmark only includes English tasks. Dealing with multilingual corpora is a real-world challenge for many NLP systems, especially for those deployed in countries where there are multiple national languages. To fully capture the distribution of real-world tasks, additional languages will be needed.

\textbf{Possible biases in data collection.} While we attempted to execute our dataset selection process as described in Section \ref{sec:dataset-description} in an unbiased manner, the datasets we ended up selecting are part of a subjective human process that may be subject to biases. For example, the organizations we work with are disproportionately in technology and policy.

\subsection{Impact}

\textbf{Offensive content.} By including a hate-speech detection dataset, we include offensive content and may harm readers of the dataset. We believe the advantages from studying hate-speech detection are likely greater than the disadvantages of publicizing hate-speech datasets.

\textbf{Prohibitive costs.} The models best equipped to perform well on RAFT will often be the massive transformer models trained by private corporations. In advancing this benchmark as a means of evaluating models, we risk further widening the gap between what a dedicated individual or team can do, and what can only be done by industry research labs with sufficient funding.

\subsection{Future Work}

\textbf{Stronger human baselines.} Human baselines are intended to tell us how well the dataset would be labeled in the absence of automated systems. For many \name{} datasets, this process would involve a stronger baseline than is easily available via a crowd-worker platform: for example, the \textit{Over} dataset would be labeled by someone with law expertise. In addition to ML submissions, we welcome efforts to collect stronger human baselines for \name{}.

\textbf{Additional automatic baselines.} We expect that systems that use prompt-based fine-tuning rather than in-context learning may provide an even stronger automatic baseline. We further expect that models that leverage the open-domain information retrieval option can surpass models that don't.

\textbf{Application-specific metrics.} Different applications care about different metrics; e.g., in some applications it is more important to minimize false positives, whereas in others the focus is on false negatives. An ideal measure of real-world value would take that into account. 

\textbf{Learning from natural language} In this work, we focused on instructions as a supplement to labeled examples. Similarly to \citet{mishra2021natural}, we found that including task-specific instructions improved performance. Like humans, NLP systems could also learn from other types of natural language. For example, could including explanations with each labeled example be used to further improve few-shot performance?

\section{Conclusion}

\name{} is a benchmark that tests language models across multiple domains on economically valuable classification tasks in the true few-shot setting. To our knowledge, this is the first multi-task benchmark designed to closely mirror how models are applied in both the task distribution and the evaluation setup. By complementing existing synthetic benchmarks designed to highlight where models fall short, it helps measure the gap between research and practice, incentivizes work that is valuable for deployed systems, and provides a template for future benchmarks that mirror deployment.

\begin{ack}

Our automatic baseline collection was subsidized by compute credits generously provided by OpenAI. Ethan Perez, Samuel Bowman, and Long Ouyang gave feedback on early versions of the \name{} concept and dataset lists. Douwe Kiela and Stella Biderman offered helpful advice on the project direction. Ross Gruetzemacher suggested inclusion of the Twitter Complaints dataset.  We thank Thomas Wolf and Simon Brandeis for discussions and advice around the design of the benchmark’s infrastructure.

\end{ack}

\bibliographystyle{plainnat}
\bibliography{references}

\newpage
\appendix

\section{Appendix}

\subsection{Dataset licensing}

The curating authors (NA, EL, and AS) bear all responsibility in the case of violation of rights. Below we provide information on the license for each dataset:

\textbf{ADE Corpus V2 (\textit{ADE}).} Unlicensed.

\textbf{Banking77 (\textit{B77}).} \href{https://github.com/PolyAI-LDN/task-specific-datasets/blob/master/LICENSE}{Creative Commons Attribution 4.0 International}.

\textbf{NeurIPS impact statement risks (\textit{NIS}).} The NeurIPS impact statement dataset has an \href{https://github.com/paulsedille/NeurIPS-Broader-Impact-Statements/blob/main/LICENSE}{MIT License}. We license the derivative NeurIPS impact statement risks dataset under \href{https://creativecommons.org/licenses/by/4.0/legalcode}{Creative Commons Attribution 4.0 International}.

\textbf{OneStopEnglish (\textit{OSE}).} \href{https://github.com/nishkalavallabhi/OneStopEnglishCorpus/blob/master/LICENSE.markdown}{Creative Commons Attribution-ShareAlike 4.0 International}.

\textbf{Overruling (\textit{Over}).} Unlicensed.

\textbf{Semiconductor org types (\textit{SOT}).} We license it under \href{https://creativecommons.org/licenses/by-nc/4.0/legalcode}{Creative Commons Attribution-NonCommercial 4.0 International}.

\textbf{Systematic review inclusion (\textit{SRI}).} \href{https://creativecommons.org/licenses/by-nc/4.0/legalcode}{Creative Commons Attribution 4.0 International}

\textbf{TAI safety research (\textit{TAI}).} \href{https://creativecommons.org/licenses/by-sa/4.0/legalcode}{Creative Commons Attribution-ShareAlike 4.0 International}

\textbf{Terms of Service (\textit{ToS}).} Unlicensed.

\textbf{TweetEval Hate (\textit{TEH}).} Unlicensed.

\textbf{Twitter complaints (\textit{TC}).} Unlicensed.

\subsection{Dataset examples}

See Table \ref{tab:examples} for one training example from each dataset.

\begin{table}[ht]

\caption{A training example from every dataset, with textual label}
\centering
\begin{tabular}{m{7em}l}
\toprule
\textbf{Dataset} & \textbf{Training Sample} \\
\midrule
ADE Corpus V2 (\textit{ADE}) & \begin{lstlisting}
{'Sentence': 'No regional side effects were noted.', 
 'Label': 'not ADE-related'}
\end{lstlisting} \\
Banking77 (\textit{B77}) & \begin{lstlisting}
{'Query': 'Is it possible for me to change my P...',
 'Label': 'change_pin'}
\end{lstlisting} \\
NeurIPS impact statement risks (\textit{NIS}) & \begin{lstlisting}
{'Paper title': 'Auto-Panoptic: Cooperative Mul...',
 'Paper link': 'https://proceedings.neurips.cc/...',
 'Impact statement': 'This work makes the first...',
 'ID': '0',
 'Label': "doesn't mention a harmful application"}
\end{lstlisting} \\
OneStopEnglish (\textit{OSE}) & \begin{lstlisting}
{'Article': 'For 85 years, it was just a grey b...', 
 'Label': 'intermediate'}
\end{lstlisting} \\
Overruling (\textit{Over}) & \begin{lstlisting}
{'Sentence': 'in light of both our holding toda...',
 'Label': 'overruling'}
\end{lstlisting} \\
Semiconductor org types (\textit{SOT}) & \begin{lstlisting}
{'Paper title': '3Gb/s AC-coupled chip-to-chip ...', 
 'Organization name': 'North Carolina State Uni...', 
 'Label': 'university'}
\end{lstlisting} \\
Systematic review inclusion (\textit{SRI}) & \begin{lstlisting}
{'Title': 'Prototyping and transforming facial ...',
 'Abstract': 'Wavelet based methods for prototy...',
 'Authors': 'Tiddeman, B.; Burt, M.; Perrett, D.',
 'Journal': 'IEEE Comput Graphics Appl',
 'Label': 'not included'}
\end{lstlisting} \\
TAI safety research (\textit{TAI}) & \begin{lstlisting}
{'Title': 'Malign generalization without intern...',
 'Abstract Note': "In my last post, I challenge...",
 'Url': 'https://www.alignmentforum.org/posts/y...',
 'Publication Year': '2020',
 'Item Type': 'blogPost',
 'Author': 'Barnett, Matthew',
 'Publication Title': 'AI Alignment Forum',
 'Label': 'TAI safety research'}
\end{lstlisting} \\
Terms of Service (\textit{ToS}) & \begin{lstlisting}
{'Sentence': 'Crowdtangle may change these term...',
 'Label': 'potentially unfair'}
\end{lstlisting} \\
TweetEval Hate (\textit{TEH}) & \begin{lstlisting}
{'Tweet': 'New to Twitter-- any men on here kno...', 
 'Label': 'not hate speech'}
\end{lstlisting} \\\\
Twitter complaints (\textit{TC}) & \begin{lstlisting}
{'Tweet text': '@HMRCcustomers No this is my fi...', 
 'Label': 'no complaint'}
\end{lstlisting} \\
\bottomrule
\end{tabular}

\label{tab:examples}
\end{table}

\subsection{Task-specific instructions}
\label{sec:task-specific-instructions}

Table \ref{tab:instruction-excerpts} contains an excerpt from the instructions for each dataset.

\begin{table}[ht]
\centering
\begin{tabular}{m{8em}m{30em}}
\toprule
\textbf{Dataset Name} & Instructions excerpt \\
\midrule
ADE Corpus V2 (\textit{ADE}) & Label the sentence based on whether it is related to an adverse drug effect (ADE). \\ 
Banking77 (\textit{B77}) & The following is a banking customer service query. \\    
NeurIPS impact statement risks (\textit{NIS}) & Label the impact statement as "mentions a harmful application" or "doesn't mention a harmful application"  based on whether it mentions a harmful application of the research done in the paper. \\ 
OneStopEnglish (\textit{OSE}) & The following is an article sourced from The Guardian newspaper, and rewritten by teachers to suit three levels of adult English as Second Language (ESL) learners: elementary, intermediate, and advanced. \\ 
Overruling (\textit{Over}) & In law, an overruling sentence is a statement that nullifies a previous case decision as a precedent, by a constitutionally valid statute or a decision by the same or higher ranking court which establishes a different rule on the point of law involved.\\ 
Semiconductor org types (\textit{SOT}) & The dataset is a list of institutions that have contributed papers to semiconductor conferences in the last 25 years, as catalogued by IEEE and sampled randomly. \\ 
Systematic review inclusion (\textit{SRI}) & Identify whether this paper should be included in a meta-review which includes the findings of systematic reviews on interventions designed to promote charitable donations. \\ 
TAI safety research (\textit{TAI}) & The contents of the paper are directly motivated by, and substantively inform, the challenge of ensuring good outcomes for Transformative AI. \\ 
Terms of Service (\textit{ToS}) & According to art.\ 3 of the Directive 93/13 on Unfair Terms in Consumer Contracts, a contractual term is unfair if: 1) it has not been individually negotiated; and 2) contrary to the requirement of good faith, it causes a significant imbalance in the parties rights and obligations, to the detriment of the consumer. \\ 
TweetEval Hate (\textit{TEH}) & Label whether the following tweet contains hate speech against either immigrants or women. \\ 
Twitter complaints (\textit{TC}) & A complaint presents a state of affairs which breaches the writer’s favorable expectation. \\ 
\bottomrule
\end{tabular}
\caption{Instructions excerpt for each dataset.}
\label{tab:instruction-excerpts}
\end{table}

Below we provide the full instructions given to human annotators and adapted for automatic baselines for each \name{} task.

\textbf{ADE Corpus V2 (\textit{ADE})}

Label the sentence based on whether it is related to an adverse drug effect (ADE). Details are described below: 

\textbf{Drugs}: Names of drugs and chemicals that include brand names, trivial names, abbreviations and systematic names were annotated. Mentions of drugs or chemicals should strictly be in a therapeutic context. This category does not include the names of metabolites, reaction byproducts, or hospital chemicals (e.g. surgical equipment disinfectants). 

\textbf{Adverse effect}: Mentions of adverse effects include signs, symptoms, diseases, disorders, acquired abnormalities, deficiencies, organ damage or death that strictly occur as a consequence of drug intake. \\

\textbf{Banking77 (\textit{B77})}

The following is a banking customer service query. Classify the query into one of the 77 categories available. \\    

\textbf{NeurIPS impact statement risks (\textit{NIS})}

Label the impact statement based on whether it mentions a harmful application of the research done in the paper. Make sure the statement is sufficient to conclude there are harmful applications of the research being done, not a past risk that this research is solving. \\

\textbf{OneStopEnglish (\textit{OSE})}

The following is an article sourced from The Guardian newspaper, and rewritten by teachers to suit three levels of adult English as Second Language (ESL) learners: elementary, intermediate, and advanced. Predict the level of the article. \\

\textbf{Overruling (\textit{Over})}

In law, an overruling sentence is a statement that nullifies a previous case decision as a precedent, by a constitutionally valid statute or a decision by the same or higher ranking court which establishes a different rule on the point of law involved. Label the sentence based on whether it is overruling or not. \\

\textbf{Semiconductor org types (\textit{SOT})} 

The dataset is a list of institutions that have contributed papers to semiconductor conferences in the last 25 years, as catalogued by IEEE and sampled randomly. The goal is to classify the institutions into one of three categories: ``university", ``company" or ``research institute". \\

\textbf{Systematic review inclusion (\textit{SRI})} 

Identify whether this paper should be included in a meta-review which includes the findings of systematic reviews on interventions designed to promote charitable donations.

Papers should be included if they meet \textbf{all} of these criteria:
\begin{enumerate}
    \item systematic reviews, scoping reviews, or similar reproducible reviews;
    \item reviews describing monetary charitable donations;
    \item reviews assessing any population of participants in any context; and
    \item peer reviewed and written in English (due to logistical constraints).
\end{enumerate}

They shouldn’t be included if they meet \textbf{any} of these criteria:
\begin{enumerate}
    \item primary research reporting new data (e.g., randomised experiments);
    \item non-systematic reviews, theory papers, or narrative reviews;
    \item reviews on cause-related marketing; and
    \item reviews of other kinds of prosocial behaviour (e.g., honesty, non-financial donations like volunteering, blood, or organ donations).
\end{enumerate}

\textbf{TAI safety research (\textit{TAI})}

Transformative AI (TAI) is defined as AI that precipitates a transition comparable to (or more significant than) the agricultural or industrial revolution.

Label a paper as ``TAI safety research" if:

\begin{enumerate}
    \item \textbf{The contents of the paper are directly motivated by, and substantively inform, the challenge of ensuring good outcomes for TAI.} The paper need not mention TAI explicitly, but it must be motivated by it, since there are far too many papers that are merely relevant to safety. Judging motivation is, unfortunately, inherently subjective, but this is necessary to avoid penalizing papers that do not explicitly mention TAI for appearance reasons, while also not including every paper on, e.g., adversarial examples (which are motivated by capabilities and near-term safety). If the paper would likely have been written even in the absence of TAI-safety concerns, it is excluded. Ultimately, we want to support researchers who are motivated by TAI safety and allow them to find each other's work
    \item \textbf{There is substantive content on AI safety, not just AI capabilities.} That said, for more speculative papers it is harder to distinguish between safety vs. not safety, and between technical vs. meta, and we err on the side of inclusion. Articles on the safety of autonomous vehicles are generally excluded, but articles on the foundations of decision theory for AGI are generally included.
    \item \textbf{The intended audience is the community of researchers.} Popular articles and books are excluded. Papers that are widely released but nevertheless have substantial research content (e.g., Bostrom's Superintelligence) are included, but papers that merely try to recruit researchers are excluded.
    \item \textbf{It meets a subjective threshold of seriousness/quality.} This is intended to be a very low threshold, and would, for instance, include anything that was accepted to be placed on the ArXiv. Web content not intended for review (e.g., blog posts) is only accepted if it has reached some (inevitably subjective) threshold of notability in the community. It is of course infeasible for us to document all blog posts that are about TAI safety, but we do not want to exclude some posts that have been influential but have never been published formally.
    \item \textbf{Peer review is not required. White papers, preprints, and book chapters are all included.}
\end{enumerate}

Otherwise, label it as ``not TAI safety research". \\

\textbf{Terms of Service (\textit{ToS})}

Label the sentence from a Terms of Service based on whether it is potentially unfair. If it seems clearly unfair, mark it as potentially unfair.

According to art. 3 of the Directive 93/13 on Unfair Terms in Consumer Contracts, a contractual term is unfair if: 1) it has not been individually negotiated; and 2) contrary to the requirement of good faith, it causes a significant imbalance in the parties rights and obligations, to the detriment of the consumer. 

Details on types of potentially unfair clauses are found below:

The \textbf{jurisdiction} clause stipulates what courts will have the competence to adjudicate disputes under the contract. Jurisdiction clauses giving consumers a right to bring disputes in their place of residence were marked as clearly fair, whereas clauses stating that any judicial proceeding takes a residence away (i.e. in a different city, different country) were marked as clearly unfair.

The \textbf{choice of law} clause specifies what law will govern the contract, meaning also what law will be applied in potential adjudication of a dispute arising under the contract. Clauses defining the applicable law as the law of the consumer’s country of residence were marked as clearly fair. In every other case, the choice of law clause was considered as potentially unfair.

The \textbf{limitation of liability} clause stipulates that the duty to pay damages is limited or excluded, for certain kind of losses, under certain conditions. Clauses that explicitly affirm non-excludable providers’ liabilities were marked as clearly fair. Clauses that reduce, limit, or exclude the liability of the service provider were marked as potentially unfair when concerning broad categories of losses or causes of them, such as any harm to the computer system because of malware or loss of data or the suspension, modification, discontinuance or lack of the availability of the service. Also those liability limitation clauses containing a blanket phrase like “to the fullest extent permissible by law”, were considered potentially unfair. Clause meant to reduce, limit, or exclude the liability of the service provider for physical injuries, intentional damages as well as in case of gross negligence were marked as clearly unfair.

The \textbf{unilateral change} clause specifies the conditions under which the service provider could amend and modify the terms of service and/or the service itself. Such clause was always considered as potentially unfair.

The \textbf{unilateral termination} clause gives provider the right to suspend and/or terminate the service and/or the contract, and sometimes details the circumstances under which the provider claims to have a right to do so.

The \textbf{contract by using} clause stipulates that the consumer is bound by the terms of use of a specific service, simply by using the service, without even being required to mark that he or she has read and accepted them. We always marked such clauses as potentially unfair.

The \textbf{content removal} gives the provider a right to modify/delete user’s content, including in-app purchases, and sometimes specifies the conditions under which the service provider may do so.

The \textbf{arbitration} clause requires or allows the parties to resolve their disputes through an arbitration process, before the case could go to court. It is therefore considered a kind of forum selection clause. However, such a clause may or may not specify that arbitration should occur within a specific jurisdiction. Clauses stipulating that the arbitration should (1) take place in a state other then the state of consumer’s residence and/or (2) be based not on law but on arbiter’s discretion were marked as clearly unfair. Clauses defining arbitration as fully optional would have to be marked as clearly fair. \\

\textbf{TweetEval Hate (\textit{TEH})} 

\textbf{WARNING: This task involves labeling offensive and hateful content, particularly toward immigrants and women.}

Label whether the following tweet contains hate speech against either immigrants or women. 

Label whether the following tweet contains hate speech against either immigrants or women. Hate Speech (HS) is commonly defined as any communication that disparages a person or a group on the basis of some characteristic such as race, color, ethnicity, gender, sexual orientation, nationality, religion, or other characteristics. Detailed guidelines are provided below, please read before labeling.

\textbf{Hate speech against immigrants}

More specifically, HS against immigrants may include:
\begin{itemize}
    \item insults, threats, denigrating or hateful expressions
    \item incitement to hatred, violence or violation of rights to individuals or groups perceived as different for somatic traits (e.g. skin color), origin, cultural traits, language, etc.
     \item presumed association of origin/ethnicity with cognitive abilities, propensity to crime, laziness or other vices
     \item references to the alleged inferiority (or superiority) of some ethnic groups with respect to others
     \item delegitimization of social position or credibility based on origin/ethnicity 
     \item references to certain backgrounds/ethnicities as a threat to the national security or welfare or as competitors in the distribution of government resources
     \item dehumanization or association with animals or entities considered inferior
\end{itemize}

While answering the question ``Is this tweet hateful?",  you must take into account the following aspects:

\begin{enumerate}
    \item the tweet content MUST have IMMIGRANTS/REFUGEES as main TARGET, or even a single individual, but considered for his/her membership in that category (and NOT for the individual characteristics) 
    \item we must deal with a message that spreads, incites, promotes or justifies HATRED OR VIOLENCE TOWARDS THE TARGET, or a message that aims at dehumanizing, hurting or intimidating the target
\end{enumerate}

The joint presence of both elements in a tweet is considered essential to determine whether the tweet has hateful contents, therefore if both of them occur, your answer will be 'Yes'.

In case even just one of these conditions is not detected, HS (at least against immigrants) is assumed not to occur, then your answer will be 'No'.

Here a list of other aspects that are NOT considered hate speech for our purposes:

\begin{itemize}
\item HATE SPEECH AGAINST OTHER TARGETS
\item offensive language
\item blasphemy
\item historical denial
\item overt incitement to terrorism
\item offense towards public servants and police officers
\item defamation
\end{itemize}

\textbf{Hate speech against women}

Label the tweet as hate speech if it is misogynous against women. A tweet is misogynous if it expresses hating towards women in particular (in the form of insulting, sexual harassment, threats of violence, stereotype, objectification and negation of male responsibility). \\

\textbf{Twitter complaints (\textit{TC})} 

A complaint presents a state of affairs which breaches the writer’s favorable expectation. Label the tweet text based on whether it contains a complaint.

\subsection{Dataset documentation}

We provide documentation using applicable questions from the datasheets framework \citep{gebru2020datasheets} for the \textit{NIS}, \textit{SOT}, and \textit{TAI} datasets. For documentation on other datasets we refer readers to the works in which the datasets were originally introduced as cited in Section \ref{sec:dataset-description}.

\subsubsection{NeurIPS impact statement risks}

The labeling section of this documentation contains information on how the impact statements were annotated based on whether they mention a harmful application. The other sections largely contain information on how the original dataset of NeurIPS impact statements  \citep{ashurst2021aiethics} was collected.

\textbf{Motivation}

\begin{itemize}
    \item \textbf{For what purpose was the dataset created? Was there a specific task in mind? Was there a specific gap that needed to be filled? Please provide a description.} The original dataset was created to evaluate the then new requirement for authors to include an "impact statement" in their 2020 NeurIPS papers. Had it been successful? What kind of things did authors mention the most? How long were impact statements on average? See \citep{ashurst2021aiethics} for more details.
    \item \textbf{Who created the dataset (e.g., which team, research group) and on behalf of which entity (e.g., company, institution, organization)?} The original dataset was created as part of a project based at the Centre for the Governance of AI, which involved individual researchers and developers from the University of Oxford, Oxford Internet Institute, Harvard Kennedy School and the Alan Turing Institute. 
    \item \textbf{Who funded the creation of dataset? If there is an associated grant, please provide the name of the grantor and the grant name and number.} The project was based at the Centre for the Governance of AI. There was no grant associated with the project. Individuals were funded by their respective organisations, or as contractors. 
\end{itemize}

\textbf{Composition}

\begin{itemize}
    \item \textbf{Is any information missing from individual instances in the dataset? If so, please provide a description, explaining why this information is missing (e.g., because it was unavailable). This does not include intentionally removed information, but might include, e.g., redacted text.} No.
    \item \textbf{Are there any errors, sources of noise, or redundancies in the dataset? If so, please provide a description.} This dataset has limitations that should be taken into consideration when using it. In particular, the method used to collect broader impact statements involved automated downloads, conversions and scraping and was not error-proof (see \href{https://github.com/paulsedille/NeurIPS-Broader-Impact-Statements/blob/main/main-dataset/notes-on-data.md}{https://github.com/paulsedille/NeurIPS-Broader-Impact-Statements/blob/main/main-dataset/notes-on-data.md} for details). Although care has been taken to identify and correct as many errors as possible, not all texts have been reviewed by a human. This means it is possible some of the broader impact statements contained in the dataset are truncated or otherwise incorrectly extracted from their original article. The original dataset also contains labels describing whether authors chose to effectively ``opt-out'' of the requirement (for example by stating that a broader impact section is ``Not Applicable''). Several statements were ambiguous in this respect, and so this label represents a subjective judgement on what constituted an opt-out. The labeling performed for this paper (whether a harmful application is mentioned) also constitutes a subjective judgment, and will contain human biases. Please see the section on Preprocessing, Cleaning, Labeling for more details. 
    \item \textbf{Does the dataset contain data that might be considered confidential (e.g., data that is protected by legal privilege or by doctor-patient confidentiality, data that includes the content of individuals’ non-public communications)? If so, please provide a description.} The dataset contains authors' names. These were scraped from publicly available scientific papers submitted to NeurIPS 2020.
    \item \textbf{Does the dataset contain data that, if viewed directly, might be offensive, insulting, threatening, or might otherwise cause anxiety? If so, please describe why.} No.
    \item \textbf{Does the dataset relate to people?} The dataset does not relate to people directly, although it does contain authors' names. These were scraped from publicly available scientific papers submitted to NeurIPS 2020.
\end{itemize}

\textbf{Collection}

\begin{itemize}
    \item \textbf{How was the data associated with each instance acquired? Was the data directly observable (e.g., raw text, movie ratings), reported by subjects (e.g., survey responses), or indirectly inferred/derived from other data (e.g., part-of-speech tags, model-based guesses for age or language)? If data was reported by subjects or indirectly inferred/derived from other data, was the data validated/verified? If so, please describe how.} The data was directly observable (raw text scraped) for the most part; although some data was taken from previous datasets (which themselves had taken it from raw text). The data was validated, but only in part, by human reviewers. Further details can be found here: https://github.com/paulsedille/NeurIPS-Broader-Impact-Statements/blob/main/main-dataset/notes-on-data.md
    \item \textbf{What mechanisms or procedures were used to collect the data (e.g., hardware apparatus or sensor, manual human curation, software program, software API)? How were these mechanisms or procedures validated?} The main dataset was collected using software, and a combination of code iteration and human review was used to validate the results. Further details may be found here: \href{https://github.com/paulsedille/NeurIPS-Broader-Impact-Statements/blob/main/main-dataset/notes-on-data.md}{https://github.com/paulsedille/NeurIPS-Broader-Impact-Statements/blob/main/main-dataset/notes-on-data.md}.
    \item \textbf{If the dataset is a sample from a larger set, what was the sampling strategy (e.g., deterministic, probabilistic with specific sampling probabilities)?} The subset annotated based on harmful applications was sampled randomly.
    \item \textbf{Who was involved in the data collection process (e.g., students, crowdworkers, contractors) and how were they compensated (e.g., how much were crowdworkers paid)?} The original dataset was created as part of a project based at the Centre for the Governance of AI, which involved individual researchers and developers as described above. The labeling for this paper (whether a harmful application is mentioned) was performed by Ought contractors. 
    \item \textbf{Does the dataset relate to people?} The dataset does not relate to people directly, although it does contain authors' names. These were scraped from publicly available scientific papers submitted to NeurIPS 2020.
    \item \textbf{Did you collect the data from the individuals in question directly, or obtain it via third parties or other sources (e.g., websites)?} The impact statements were collected from the NeurIPS websites. Metadata included in the original dataset was collected from the NeurIPS chairs, and websites (for example where affiliated institutions are geographically based). See \citep{ashurst2021aiethics} for further details. The labeling for this paper (whether a harmful application is mentioned) was collected from the contractors directly.
\end{itemize}

\textbf{Preprocessing, Cleaning, Labeling}

\begin{itemize}
    \item \textbf{Was any preprocessing/cleaning/labeling of the data done (e.g., discretization or bucketing, tokenization, part-of-speech tagging, SIFT feature extraction, removal of instances, processing of missing values)?} For the original dataset \citep{ashurst2021aiethics}, the manuscript pdfs for accepted papers were obtained from the NeurIPS 2020 proceedings website. The pdfs were converted to XML, and  
the title and impact statement section were extracted. The dataset was appended with information about paper subject area, author names,  affiliations, affiliation type and affiliation institution locations, as follows. Primary and secondary subject area, as selected by authors on submission, were supplied to us by the NeurIPS programme chairs. Author names and affiliations were obtained from separate scrapes of the NeurIPS papers. Each affiliation was tagged with a location and type (industry or academia) based on \citep{neurips_2020_locations} and \citep{affiliation_type_google_sheet} respectively. Further details on the generation of the original dataset, and its assumptions and limitations, can be found at \href{https://github.com/paulsedille/NeurIPS-Broader-Impact-Statements/blob/main/main-dataset/notes-on-data.md}{https://github.com/paulsedille/NeurIPS-Broader-Impact-Statements/blob/main/main-dataset/notes-on-data.md}. Contractors paid by Ought performed the labeling of whether impact statements mention harmful applications. A majority vote was taken from three annotators.
    \item \textbf{Was the “raw” data saved in addition to the preprocessed/cleaned/labeled data (e.g., to support unanticipated future uses)?} The original NeurIPS impact statements data is available at \href{https://github.com/paulsedille/NeurIPS-Broader-Impact-Statements}{https://github.com/paulsedille/NeurIPS-Broader-Impact-Statements}. The accepted papers containing the statements can also be found at \href{https://proceedings.neurips.cc/paper/2020}{https://proceedings.neurips.cc/paper/2020}.
\end{itemize}

\textbf{Uses}

\begin{itemize}
    \item \textbf{Has the dataset been used for any tasks already? If so, please provide a description.} An analysis of the original dataset has been prepared by the dataset authors, which can be found in \citet{ashurst2021aiethics}.
    \item \textbf{What (other) tasks could the dataset be used for?} Other researchers are encouraged to use the dataset to provide further analysis on the outcomes of the NeurIPS broader impact requirement. The dataset could also be used for additional meta-analysis of NeurIPS 2020 accepted papers.
    \item \textbf{Is there anything about the composition of the dataset or the way it was collected and preprocessed/cleaned/labeled that might impact future uses? For example, is there anything that a future user might need to know to avoid uses that could result in unfair treatment of individuals or groups (e.g., stereotyping, quality of service issues) or other undesirable harms (e.g., financial harms, legal risks) If so, please provide a description. Is there anything a future user could do to mitigate these undesirable harms?} This dataset has limitations that should be taken into consideration when using it. In particular, the method used to collect broader impact statements involved automated downloads, conversions and scraping and was not error-proof. Although care has been taken to identify and correct as many errors as possible, not all texts have been reviewed by a human. This means it is possible some of the broader impact statements contained in the dataset are truncated or otherwise incorrectly extracted from their original article. More details may be found at \href{https://github.com/paulsedille/NeurIPS-Broader-Impact-Statements/blob/main/main-dataset/notes-on-data.md}{https://github.com/paulsedille/NeurIPS-Broader-Impact-Statements/blob/main/main-dataset/notes-on-data.md}. For this paper, individual labelers were asked whether harmful applications were mentioned in the statement, but what constitutes a harmful application is of course highly subjective, and will depend on the particular views and experiences of the labeler. For example, many applications will provide some benefits to some individuals and groups, while creating risks and harms to others. The intention was to capture a rough measure of whether the authors had intended to point out potential negative effects that could arise from the use of their work, or whether they chose to limit to potential positive impacts only. This will likely exclude applications that are typically viewed as beneficial or neutral, despite the fact that such applications can cause harm to individuals or subgroups in society. We therefore urge caution in how such labels are interpreted for future tasks. 
\end{itemize}

\subsubsection{Semiconductor org types}

This Labeling section of this documentation contains information on how the semiconductor organizations were annotated by type. The other sections mainly contain information describing how the unlabeled dataset of semiconductor organizations was collected.

\textbf{Motivation}

\begin{itemize}
    \item \textbf{For what purpose was the dataset created? Was there a specific task in mind? Was there a specific gap that needed to be filled? Please provide a description.} The data set was originally created to understand better which countries’ organisations have contributed most to semiconductor R\&D over the past 25 years using three main conferences. Moreover, to estimate the share of academic and private sector contributions, the organisations were classified as “university”, “research institute” or “company”.
    \item \textbf{Who created the dataset (e.g., which team, research group) and on behalf of which entity (e.g., company, institution, organization)?} The data science unit of Stiftung Neue Verantwortung (Berlin).
    \item \textbf{Who funded the creation of dataset? If there is an associated grant, please provide the name of the grantor and the grant name and number.} The Stiftung Mercator is funding the data science unit in general

\end{itemize}

\textbf{Composition}

\begin{itemize}
    \item \textbf{Is any information missing from individual instances in the dataset? If so, please provide a description, explaining why this information is missing (e.g., because it was unavailable). This does not include intentionally removed information, but might include, e.g., redacted text.} This data set is a sample of 500 out of many more organisations. Examples where the institution names contain “universit” were deleted because all language models can classify this as "university" and no discrimination is gained.
    \item \textbf{Are there any errors, sources of noise, or redundancies in the dataset? If so, please provide a description.} The human-created labels could be wrong.
    \item \textbf{Does the dataset contain data that might be considered confidential (e.g., data that is protected by legal privilege or by doctor-patient confidentiality, data that includes the content of individuals’ non-public communications)? If so, please provide a description.} No.
    \item \textbf{Does the dataset contain data that, if viewed directly, might be offensive, insulting, threatening, or might otherwise cause anxiety? If so, please describe why.} No.
    \item \textbf{Does the dataset relate to people?} No.
\end{itemize}

\textbf{Collection}

\begin{itemize}
    \item \textbf{What mechanisms or procedures were used to collect the data (e.g., hardware apparatus or sensor, manual human curation, software program, software API)? How were these mechanisms or procedures validated?} We used the IEEE API to obtain institutions that contributed papers to semiconductor conferences in the last 25 years. This is a random sample of 500 of them with a corresponding conference paper title. The three conferences were the International Solid-State Circuits Conference (ISSCC), the Symposia on VLSI Technology and Circuits (VLSI) and the International Electron Devices Meeting (IEDM).
    \item \textbf{If the dataset is a sample from a larger set, what was the sampling strategy (e.g., deterministic, probabilistic with specific sampling probabilities)?} It was probabilistic. Duplicate entries (by organisation name) were deleted.
    \item \textbf{Who was involved in the data collection process (e.g., students, crowdworkers, contractors) and how were they compensated (e.g., how much were crowdworkers paid)?} A student was involved and paid according to German law.
    \item \textbf{Over what timeframe was the data collected? Does this timeframe match the creation timeframe of the data associated with the instances (e.g., recent crawl of old news articles)? If not, please describe the timeframe in which the data associated with the instances was created.} March 2021
\end{itemize}

\textbf{Preprocessing, Cleaning, Labeling}

\begin{itemize}
    \item \textbf{Was any preprocessing/cleaning/labeling of the data done (e.g., discretization or bucketing, tokenization, part-of-speech tagging, SIFT feature extraction, removal of instances, processing of missing values)?} Yes. Contractors paid by Ought performed the labeling of organization types. A majority vote was taken from 3 annotators.
\end{itemize}

\textbf{Distribution}

\begin{itemize}
    \item \textbf{Will the dataset be distributed under a copyright or other intellectual property (IP) license, and/or under applicable terms of use (ToU)? If so, please describe this license and/or ToU, and provide a link or other access point to, or otherwise reproduce, any relevant licensing terms or ToU, as well as any fees associated with these restrictions.} It can only be used for non-commercial research purposes. See \href{https://developer.ieee.org/docs/read/API_Use_Cases_Examples}{here} and \href{https://developer.ieee.org/docs/read/IEEE_Xplore_Metadata_API_Overview}{here}. The annotated data is licensed under \href{https://creativecommons.org/licenses/by-nc/4.0/legalcode}{Creative Commons Attribution-NonCommercial 4.0 International}. 
\end{itemize}

\subsubsection{TAI Safety Research}

\textbf{Motivation}

\begin{itemize}
    \item \textbf{For what purpose was the dataset created? Was there a specific task in mind? Was there a specific gap that needed to be filled? Please provide a description.} The primary motivations for assembling this database were to: (1) Aid potential donors in assessing organizations focusing on TAI safety by collecting and analyzing their research output. (2) Assemble a comprehensive bibliographic database that can be used as a base for future projects, such as a living review of the field.
    \item \textbf{Who created the dataset (e.g., which team, research group) and on behalf of which entity (e.g., company, institution, organization)?} Angelica Deibel and myself (Jess Riedel).  We did not do it on behalf of any entity.
    \item \textbf{Who funded the creation of dataset? If there is an associated grant, please provide the name of the grantor and the grant name and number.} I volunteered my own time and paid Angelica Deibel for her time from my personal funds.
\end{itemize}

\textbf{Composition}

\begin{itemize}
    \item \textbf{Is any information missing from individual instances in the dataset? If so, please provide a description, explaining why this information is missing (e.g., because it was unavailable). This does not include intentionally removed information, but might include, e.g., redacted text.} Not really sure what this means for our case.  There's no redacted information, but there are undoubtedly tons of papers we failed to find in our literature search.  Also, we kept/excluded articles based on a set of subjective criteria we invented, and we undoubtedly made mistakes applying this criteria.
    \item \textbf{Are there any errors, sources of noise, or redundancies in the dataset? If so, please provide a description.} See above.  No redundancies that I know of.
    \item \textbf{Does the dataset contain data that might be considered confidential (e.g., data that is protected by legal privilege or by doctor-patient confidentiality, data that includes the content of individuals’ non-public communications)? If so, please provide a description.} No.
    \item \textbf{Does the dataset contain data that, if viewed directly, might be offensive, insulting, threatening, or might otherwise cause anxiety? If so, please describe why.} No.
    \item \textbf{Does the dataset relate to people? If not, you may skip the remaining questions in this section.} Sort of. It's a database of papers, and those papers have authors.
    \item \textbf{Does the dataset identify any subpopulations (e.g., by age, gender)? If so, please describe how these subpopulations are identified and provide a description of their respective distributions within the dataset. Is it possible to identify individuals (i.e., one or more natural persons), either directly or indirectly (i.e., in combination with other data) from the dataset? If so, please describe how.} It's a database of papers, and those papers have authors.  This information is already public.
    \item \textbf{Does the dataset contain data that might be considered sensitive in any way (e.g., data that reveals racial or ethnic origins, sexual orientations, religious beliefs, political opinions or union memberships, or locations; financial or health data; biometric or genetic data; forms of government identification, such as social security numbers; criminal history)? If so, please provide a description.} No.
\end{itemize}

\textbf{Collection}

\begin{itemize}
    \item \textbf{How was the data associated with each instance acquired? Was the data directly observable (e.g., raw text, movie ratings), reported by subjects (e.g., survey responses), or indirectly inferred/derived from other data (e.g., part-of-speech tags, model-based guesses for age or language)? If data was reported by subjects or indirectly inferred/derived from other data, was the data validated/verified? If so, please describe how.} We asked TAI safety organizations for what their employees had written, emailed some individual authors, and searched Google Scholar. See the LessWrong post for more details: https://www.lesswrong.com/posts/4DegbDJJiMX2b3EKm/tai-safety-bibliographic-database
    \item \textbf{What mechanisms or procedures were used to collect the data (e.g., hardware apparatus or sensor, manual human curation, software program, software API)? How were these mechanisms or procedures validated?} Mostly be hand.  We collected citation information using an automated API call to Google Scholar. See the LessWrong post for more details: https://www.lesswrong.com/posts/4DegbDJJiMX2b3EKm/tai-safety-bibliographic-database
    \item \textbf{Who was involved in the data collection process (e.g., students, crowdworkers, contractors) and how were they compensated (e.g., how much were crowdworkers paid)?} It was Angelica Deibel and me.  I volunteered and paid Angelica \$20/hour.
    \item \textbf{Over what timeframe was the data collected? Does this timeframe match the creation timeframe of the data associated with the instances (e.g., recent crawl of old news articles)? If not, please describe the timeframe in which the data associated with the instances was created.} It was collected haphazardly between in 2019 and 2020
    \item \textbf{Were any ethical review processes conducted (e.g., by an institutional review board)? If so, please provide a description of these review processes, including the outcomes, as well as a link or other access point to any supporting documentation.} No.
    \item \textbf{Does the dataset relate to people?} It's a database of papers, which have authors.
    \item \textbf{Did you collect the data from the individuals in question directly, or obtain it via third parties or other sources (e.g., websites)?} Both.
    \item \textbf{Were the individuals in question notified about the data collection? If so, please describe (or show with screenshots or other information) how notice was provided, and provide a link or other access point to, or otherwise reproduce, the exact language of the notification itself.} We asked authors to suggest papers that should be included in the database.
    \item \textbf{Has an analysis of the potential impact of the dataset and its use on data subjects (e.g., a data protection impact analysis)been conducted? If so, please provide a description of this analysis, including the outcomes, as well as a link or other access point to any supporting documentation.} No.
\end{itemize}

\textbf{Preprocessing, Cleaning, Labeling}

\begin{itemize}
    \item \textbf{Was any preprocessing/cleaning/labeling of the data done (e.g., discretization or bucketing, tokenization, part-of-speech tagging, SIFT feature extraction, removal of instances, processing of missing values)?} Yes.  See the LessWrong post for more details on our labels, which was done largely by hand, on citation numbers, collected from Google Scholar by automated API call, and on the basic bibliographic information, which was collected with the automated tools from Zotero: \href{https://www.lesswrong.com/posts/4DegbDJJiMX2b3EKm/tai-safety-bibliographic-database}{https://www.lesswrong.com/posts/4DegbDJJiMX2b3EKm/tai-safety-bibliographic-database}
    \item \textbf{Was the “raw” data saved in addition to the preprocessed/cleaned/labeled data (e.g., to support unanticipated future uses)?} No.  There was no clean distinction between raw and processed data.  We used several automated tools that interacted, plus corrections and additions by hand.
    \item \textbf{Is the software used to preprocess/clean/label the instances available? If so, please provide a link or other access point.} See link to the Citation numbers API called for Google Scholar in the the LessWrong post for more details: \href{https://www.lesswrong.com/posts/4DegbDJJiMX2b3EKm/tai-safety-bibliographic-database}{https://www.lesswrong.com/posts/4DegbDJJiMX2b3EKm/tai-safety-bibliographic-database}
\end{itemize}

\textbf{Uses}

\begin{itemize}
    \item \textbf{Has the dataset been used for any tasks already? If so, please provide a description.} Yes, for the report we posted on LessWrong \href{https://www.lesswrong.com/posts/4DegbDJJiMX2b3EKm/tai-safety-bibliographic-database#Inclusion___categorization1}{here}. It was also used by "Larks" in \href{https://forum.effectivealtruism.org/posts/K7Z87me338BQT3Mcv/2020-ai-alignment-literature-review-and-charity-comparison}{his review}.
    \item \textbf{Is there a repository that links to any or all papers or systems that use the dataset? If so, please provide a link or other access point.} No, this hasn't been used in any academic papers yet.
    \item \textbf{Is there anything about the composition of the dataset or the way it was collected and preprocessed/cleaned/labeled that might impact future uses? For example, is there anything that a future user might need to know to avoid uses that could result in unfair treatment of individuals or groups (e.g., stereotyping, quality of service issues) or other undesirable harms (e.g., financial harms, legal risks) If so, please provide a description. Is there anything a future user could do to mitigate these undesirable harms?} No.
    \item \textbf{Are there tasks for which the dataset should not be used? If so, please provide a description.} Don't use it to create a dangerous AI that could bring the end of days.
\end{itemize}

\textbf{Distribution}

\begin{itemize}
    \item \textbf{Will the dataset be distributed under a copyright or other intellectual property (IP) license, and/or under applicable terms of use (ToU)? If so, please describe this license and/or ToU, and provide a link or other access point to, or otherwise reproduce, any relevant licensing terms or ToU, as well as any fees associated with these restrictions.} As mentioned in the LessWrong post: ``We release the Zotero database under the Creative Commons Attribution-ShareAlike 4.0 International License.  In short, the means you are free to use, modify, and reproduce the database for anything so long as you cite us and release any derivative works under the same license."
    \item \textbf{Have any third parties imposed IP-based or other restrictions on the data associated with the instances? If so, please describe these restrictions, and provide a link or other access point to, or otherwise reproduce, any relevant licensing terms, as well as any fees associated with these restrictions.} No. The CC-SA-BY license is the only restriction
    \item \textbf{Do any export controls or other regulatory restrictions apply to the dataset or to individual instances? If so, please describe these restrictions, and provide a link or other access point to, or otherwise reproduce, any supporting documentation.} No.
\end{itemize}

\subsection{GPT-3 baseline details}
\label{sec:auto-baseline-details}

The code for the GPT-3 baseline is available at \baselinesURL{} under an MIT license. Running the automatic baseline of GPT-3 davinci on the test sets cost approximately \$2,600.

\subsubsection{Parameter selection}
Tables \ref{tab:instruction-results}, \ref{tab:semantic-results} and \ref{tab:n-example-results} detail the results of parameter selection runs. All runs were done using GPT-3.

We mistakenly use 50 rather than 25 training examples in the prompt for \textit{TEH} when running in-context baselines, despite 25 performing better in LOOCV.

When running in-context baselines besides GPT-3, we use the same number of training examples in the prompt. Note that this may be suboptimal due to other models having smaller context windows; we leave improving upon these baselines to future work.

\begin{table}[ht]
\centering
\begin{tabular}{m{6em} M{1.5em} M{1.5em} M{1.5em} M{1.5em} M{1.5em} M{1.5em} M{1.5em} M{1.5em} M{1.5em} M{1.5em} M{1.5em} M{1.5em} M{1.5em}}
\toprule
\textbf{Instructions} & \textbf{Avg} & \textit{ADE} & \textit{B77} & \textit{NIS} & \textit{OSE} & \textit{Over} & \textit{SOT} & \textit{SRI} & \textit{TAI} & \textit{ToS} & \textit{TEH} & \textit{TC} \\  
\midrule
Task-Specific & \textbf{0.593} & 0.752 & 0.081 & 0.566 & 0.231 & 0.940 & 0.777 & 0.495 & 0.480 & 0.691 & 0.731 & 0.780 \\
Generic & \textbf{0.551} & 0.797 & 0.052 & 0.476 & 0.184 & 0.899 & 0.717 & 0.495 & 0.462 & 0.438 & 0.708 & 0.830 \\
\bottomrule
\end{tabular}
\caption{LOO Cross Validation performance for task-specific versus generic instructions, F1 scores. The experiment was run with 20 training examples for all datasets and no semantic selection.}
\label{tab:instruction-results}
\end{table}

\begin{table}[ht]
\centering
\begin{tabular}{m{6em} M{1.5em} M{1.5em} M{1.5em} M{1.5em} M{1.5em} M{1.5em} M{1.5em} M{1.5em} M{1.5em} M{1.5em} M{1.5em} M{1.5em} M{1.5em}}
\toprule
\textbf{Selection} & \textbf{Avg} & \textit{ADE} & \textit{B77} & \textit{NIS} & \textit{OSE} & \textit{Over} & \textit{SOT} & \textit{SRI} & \textit{TAI} & \textit{ToS} & \textit{TEH} & \textit{TC} \\  
\midrule
Semantic & \textbf{0.622} & 0.696 & 0.098 & 0.635 & 0.454 & 0.940 & 0.716 & 0.419 & 0.696 & 0.578 & 0.778 & 0.836 \\
Random & \textbf{0.593} & 0.752 & 0.081 & 0.566 & 0.231 & 0.94 & 0.777 & 0.495 & 0.480 & 0.691 & 0.731 & 0.780 \\
\bottomrule
\end{tabular}
\caption{LOO Cross Validation performance for semantic versus random training example selection, F1 scores. The experiment was run with 20 training examples for all datasets and task-specific instructions.}
\label{tab:semantic-results}
\end{table}

\begin{table}[ht]
\centering
\begin{tabular}{m{6em} M{1.5em} M{1.5em} M{1.5em} M{1.5em} M{1.5em} M{1.5em} M{1.5em} M{1.5em} M{1.5em} M{1.5em} M{1.5em} M{1.5em} M{1.5em}}
\toprule
\textbf{Examples} & \textbf{Avg} & \textit{ADE} & \textit{B77} & \textit{NIS} & \textit{OSE} & \textit{Over} & \textit{SOT} & \textit{SRI} & \textit{TAI} & \textit{ToS} & \textit{TEH} & \textit{TC} \\  
\midrule
5 & 0.611 & 0.696 & 0.076 & 0.571 & \textbf{0.528} & 0.860 & \textbf{0.745} & 0.474 & \textbf{0.672} & \textbf{0.789} & 0.618 & 0.688 \\
10 & 0.593 & 0.667 & \textbf{0.096} & 0.559 & 0.456 & 0.920 & 0.623 & \textbf{0.479} & 0.642 & 0.610 & 0.735 & 0.733 \\
25 & 0.617 & \textbf{0.714} & 0.090 & \textbf{0.740} & 0.445 & \textbf{0.960} & 0.591 & 0.412 & 0.643 & 0.548 & \textbf{0.778} & \textbf{0.862} \\
49 & 0.598 & 0.653 & 0.074 & 0.643 & 0.394 & 0.960 & 0.692 & 0.375 & 0.586 & 0.643 & 0.718 & 0.842 \\
\bottomrule
\end{tabular}
\caption{LOO Cross Validation performance for number of training examples, F1 scores. The experiment was run with task-specific instructions and semantic selection of training examples.}
\label{tab:n-example-results}
\end{table}

\subsection{AdaBoost baseline details}

We concatenated all non-label data in every training example into a single string, separated by periods, then constructed $n$-grams from all words and adjacent sets of $n$ words in the dataset for $n \in [1, 5]$ after removing letter cases and certain special symbols. Each training or test example was vectorized as the count of each $n$-gram in the example. For the base estimator, we used decision trees with a maximum depth of 3. We ensembled 100 estimators with a learning rate of 1.0.

We tuned several hyperparameters in our AdaBoost implementation. First, we tested the learning rate of AdaBoost, the rate at which the weights of the ensembled classifiers are changed, finding that it didn't change results substantially from within a reasonable range. We then tested a number of different depths of decision trees in the ensemble, finding that low depths were ideal. Finally, we tested the number of trees to ensemble, finding that around 50 to 100 trees perform the best. All hyperparameters were tuned with leave-one-out cross validation.

\begin{table}[ht]
\centering
\begin{tabular}{m{6em} M{1.5em} M{1.5em} M{1.5em} M{1.5em} M{1.5em} M{1.5em} M{1.5em} M{1.5em} M{1.5em} M{1.5em} M{1.5em} M{1.5em} M{1.5em}}
\toprule
\textbf{Learn Rate} & \textbf{Avg} & \textit{ADE} & \textit{B77} & \textit{NIS} & \textit{OSE} & \textit{Over} & \textit{SOT} & \textit{SRI} & \textit{TAI} & \textit{ToS} & \textit{TEH} & \textit{TC} \\  
\midrule
0.03 & \textbf{0.547} & 0.587 & 0.012 & 0.760 & 0.344 & 0.900 & 0.538 & 0.495 & 0.720 & 0.621 & 0.475 & 0.561 \\
0.1 & \textbf{0.544} & 0.587 & 0.004 & 0.760 & 0.344 & 0.900 & 0.515 & 0.495 & 0.720 & 0.621 & 0.475 & 0.561 \\
0.3 & \textbf{0.536} & 0.714 & 0.009 & 0.667 & 0.352 & 0.919 & 0.422 & 0.495 & 0.660 & 0.642 & 0.451 & 0.561 \\
1.0 & \textbf{0.548} & 0.636 & 0.000 & 0.718 & 0.444 & 0.919 & 0.385 & 0.495 & 0.699 & 0.621 & 0.538 & 0.569 \\
3.0 & \textbf{0.410} & 0.643 & 0.000 & 0.392 & 0.432 & 0.597 & 0.284 & 0.495 & 0.653 & 0.368 & 0.333 & 0.319 \\
\bottomrule
\end{tabular}
\caption{LOO Cross Validation performance for learning rate, F1 scores from an AdaBoost ensemble classifier of 50 depth-1 decision trees trained on $n$-grams of the dataset for $n \in [1, 5]$.}
\label{tab:lr-results}
\end{table}

\begin{table}[ht]
\centering
\begin{tabular}{m{6em} M{1.5em} M{1.5em} M{1.5em} M{1.5em} M{1.5em} M{1.5em} M{1.5em} M{1.5em} M{1.5em} M{1.5em} M{1.5em} M{1.5em} M{1.5em}}
\toprule
\textbf{Depth} & \textbf{Avg} & \textit{ADE} & \textit{B77} & \textit{NIS} & \textit{OSE} & \textit{Over} & \textit{SOT} & \textit{SRI} & \textit{TAI} & \textit{ToS} & \textit{TEH} & \textit{TC} \\  
\midrule
1 & \textbf{0.546} & 0.636 & 0.000 & 0.718 & 0.466 & 0.919 & 0.385 & 0.495 & 0.699 & 0.621 & 0.494 & 0.569 \\
2 & \textbf{0.511} & 0.592 & 0.000 & 0.716 & 0.405 & 0.900 & 0.366 & 0.495 & 0.466 & 0.527 & 0.626 & 0.524 \\
3 & \textbf{0.549} & 0.721 & 0.004 & 0.735 & 0.463 & 0.919 & 0.366 & 0.495 & 0.507 & 0.621 & 0.626 & 0.586 \\
4 & \textbf{0.531} & 0.556 & 0.000 & 0.672 & 0.410 & 0.880 & 0.438 & 0.495 & 0.607 & 0.602 & 0.524 & 0.653 \\
5 & \textbf{0.506} & 0.619 & 0.004 & 0.583 & 0.318 & 0.860 & 0.360 & 0.495 & 0.451 & 0.602 & 0.592 & 0.684 \\
\bottomrule
\end{tabular}
\caption{LOO Cross Validation performance for depth of trees, F1 scores from an AdaBoost ensemble classifier of 50 decision trees with learning rate 1.0 trained on $n$-grams of the dataset for $n \in [1, 5]$.}
\label{tab:depth-results}
\end{table}

\begin{table}[ht]
\centering
\begin{tabular}{m{6em} M{1.5em} M{1.5em} M{1.5em} M{1.5em} M{1.5em} M{1.5em} M{1.5em} M{1.5em} M{1.5em} M{1.5em} M{1.5em} M{1.5em} M{1.5em}}
\toprule
\textbf{\# Trees} & \textbf{Avg} & \textit{ADE} & \textit{B77} & \textit{NIS} & \textit{OSE} & \textit{Over} & \textit{SOT} & \textit{SRI} & \textit{TAI} & \textit{ToS} & \textit{TEH} & \textit{TC} \\  
\midrule
10 & \textbf{0.547} & 0.649 & 0.000 & 0.616 & 0.394 & 0.860 & 0.516 & 0.495 & 0.660 & 0.691 & 0.582 & 0.554 \\
50 & \textbf{0.544} & 0.636 & 0.000 & 0.697 & 0.482 & 0.919 & 0.381 & 0.495 & 0.699 & 0.621 & 0.491 & 0.569 \\
100 & \textbf{0.554} & 0.667 & 0.000 & 0.756 & 0.318 & 0.919 & 0.390 & 0.495 & 0.697 & 0.642 & 0.592 & 0.616 \\
500 & \textbf{0.537} & 0.649 & 0.000 & 0.814 & 0.305 & 0.919 & 0.385 & 0.495 & 0.557 & 0.642 & 0.592 & 0.548 \\
\bottomrule
\end{tabular}
\caption{LOO Cross Validation performance for number of trees, F1 scores from an AdaBoost ensemble classifier with learning rate 1.0 trained on $n$-grams of the dataset for $n \in [1, 5]$.}
\label{tab:trees-results}
\end{table}

\subsection{Human baseline details}
\label{sec:human-baseline-details}

\subsubsection{Labeling process}

For each dataset, we first conduct a qualification phase with 20 data points from the training set, showing labelers the other 30 as reference examples. Labelers who label at least 10 data points and achieved at least median accuracy advance to the annotation phase. In the annotation round, we collect 5 labels for each of the 100 data points. We then take the plurality vote for each data point, breaking ties randomly. 

Due to extreme class imbalance, we conduct only an annotation phase of 200 data points for the \textit{SRI} dataset.

\subsubsection{Instructions}

We attempted to mimic annotation instructions reported by the works introducing datasets whenever possible. The instructions we gave to annotators was as follows (parts enclosed in brackets denote variations in the instructions depending on the task or phase):

\begin{quote}
    [If qualification phase: This task will serve as a qualification stage for annotation on a larger set. Label at least 10 examples to be considered for qualification for the annotation task. Please only complete this qualification if you’re available to label 100 more data points in the next day.] \\
    
    [Task-specific instructions] \\
    
    There are 50 [30 if qualification phase] labeled examples here to help you. If it seems that the instructions and examples are in conflict, use the examples as a guide. \\

    You may use info on the internet (e.g. Google searches) to help you. \\
 
    We know that labeling accuracy will (a) vary some based on level of background knowledge and (b) have some inherent subjectivity. Please select your best guess for each data point.
\end{quote}

Task-specific instructions are detailed in Section \ref{sec:task-specific-instructions}.

\subsubsection{Costs}

We spent \$2,030 compensating crowdworkers for human baselines. We conservatively estimate that workers were paid \$15/hr.

\end{document}